IAC-22-B1.IP

# Fast Fourier Convolution Based Remote Sensor Image Object Detection for Earth Observation

Gu Lingyun[a], Eugene Popov[b], Dong Ge[c]*

[a] *Institute of Electronics and Telecommunications, Peter the Great Saint- Petersburg Polytechnic University, Saint Petersburg 195251, Russia*, gulingyun7@gmail.com
[b] *Institute of Electronics and Telecommunications, Peter the Great Saint- Petersburg Polytechnic University, Saint Petersburg 195251, Russia*, popov@spbstu.ru
[c] *School of Aerospace Engineering, Tsinghua University, Beijing 100084, China*, dongge@tsinghua.edu.cn
\* Corresponding Author

**Abstract**
Remote sensor image object detection is an important technology for Earth observation, and is used in various tasks such as forest fire monitoring and ocean monitoring. Image object detection technology, despite the significant developments, is struggling to handle remote sensor images and small-scale objects, due to the limited pixels of small objects. Numerous existing studies have demonstrated that an effective way to promote small object detection is to introduce the spatial context. Meanwhile, recent researches for image classification have shown that spectral convolution operations can perceive long-term spatial dependence more efficiently in the frequency domain than spatial domain. Inspired by this observation, we propose a Frequency-aware Feature Pyramid Framework (FFPF) for remote sensing object detection, which consists of a novel Frequency-aware ResNet (F-ResNet) and a Bilateral Spectral-aware Feature Pyramid Network (BS-FPN). Specifically, the F-ResNet is proposed to perceive the spectral context information by plugging the frequency domain convolution into each stage of the backbone, extracting richer features of small objects. To the best of our knowledge, this is the first work to introduce frequency-domain convolution into remote sensing object detection task. In addition, the BSFPN is designed to use a bilateral sampling strategy and skipping connection to better model the association of object features at different scales, towards unleashing the potential of the spectral context information from F-ResNet. Extensive experiments are conducted for object detection in the optical remote sensing image dataset (DIOR and DOTA). The experimental results demonstrate the excellent performance of our method. It achieves an average accuracy (mAP) without any tricks.

## 1. Introduction

Remote sensor (RS) image object detection is an important task in Earth observation, and it is widely studied due to its great potential in disaster prevention and environmental monitoring. Remote sensing object detection refers to locating the position of specific objects and determining their category from images taken by satellites or aircrafts.

With the development of deep learning, object detection approaches have shown excellent performance in generic scenes. Generally, deep learning-based object detection approaches are divided into two main categories: two-stage approaches and one-stage approaches. Two-stage approaches first generate candidate regions, and then perform classification and location regression [1]–[3], while one-stage approaches produce object detection results without generating any object proposals [4], [5]. With the progress of remote sensing technology, it is easier to obtain a large number of images. Therefore, data-driven deep learning approaches are potentially applied to remote sensing object detection.

Compared to generic scenes, RS scenes contain more small objects that are difficult to recognize due to the lack of sufficient details. To address this problem, numerous studies introduce the rich context into model to effectively improve the detection accuracy of small objects.

An example can explain this intuitively: a small boat alone is difficult to recognize, but with the introduction of its surrounding context (e.g., ocean, port, coast.), the detection of small boats becomes easier, as Fig. 1 shown. To perceive the context information, a straightforward approach is to adopt the convolutions with large kernels by expanding the receptive field, but it increases a significant computational cost. Existing methods also try to employ attention mechanism [6] or cross-scale fusion [7] to learn rich contexts for small object detection. In specific, YOLT [8] proposes a pass-through layer that allows the detector in accessing fine-grained features to improve the fidelity of small objects. SCRDet [9] designs a sampling fusion network that fuses multi-layer features with effective anchor sampling to improve sensitivity to small objects. ARSF [10] designs a spatial feature fusion module to enhance the features of small objects. These methods consider explicitly extracting the context information in the spatial domain.

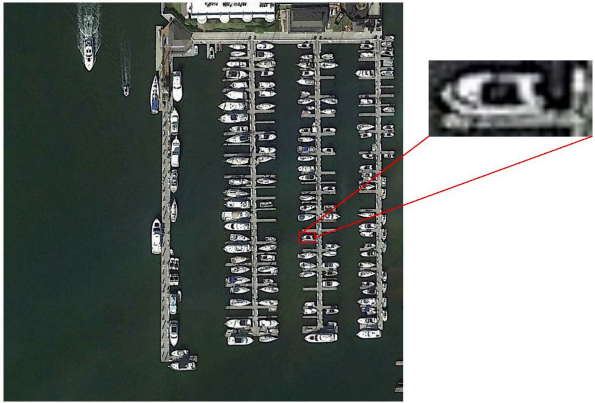

Fig. 1. The context of the small object is necessary to recognize the boat in this picture.

Recently, a series of works [11]–[13] adopt spectral neural networks to efficiently implement non-local receptive field from frequency perspective. According to the spectral convolution theorem [14], each value in the frequency domain is computed from all values in the spatial domain, and updating a value in the spectral domain can affect all spatial domain data globally. Based on this theorem, the Fast Fourier Convolution [11] was proposed to explore an efficient combination of non-local receptive fields and cross-scale fusion within the convolutional unit in the frequency domain for image recognition, keypoint detection, and action recognition.

Suvorov et al. [13] utilize Fast Fourier convolution to obtain image-level receptive fields for image inpainting task. A global filtering network [12] is proposed to learn global context in the frequency domain with log-linear complexity for image classification. Due to the advanced properties of Fourier convolution, we consider the introduction of spectral convolution into the remote sensing community to facilitate the detection of small objects, which is the first work to explore frequency domain operation on remote sensing object detection tasks.

In this paper, a novel Frequency-Aware Feature Pyramid Framework (FFPF) is proposed to solve the small object detection problem of remote sensing images effectively. FFPF integrates two novel components: Frequency-aware ResNet (F-ResNet) and a Bilateral Spectral-aware Feature Pyramid Network (BS-FPN). Specifically, the F-ResNet is proposed to perceive the spectral context information by plugging the frequency-domain convolution into each stage of the backbone, extracting richer features of small objects. BS-FPN is designed to use a bilateral sampling strategy and skipping connection to better model the association of object features at different scales, towards unleashing the potential of the spectral context information from F-ResNet.

We demonstrate the effectiveness of our approach by conducting extensive experiments on the public remote sensing object detection datasets DIOR [15] and DOTA[26]. In general, the contributions of this paper can be summarized as follows:

1) A novel Frequency-aware ResNet (F-ResNet) with the global receptive field is proposed, which consists of two parts: a spatial convolutional backbone to extract spatial features and several spectral convolutional modules named FUs to obtain spectral global context. To the best of our knowledge, this is the first work to explore frequency domain operation on remote sensing object detection tasks.

2) A Bilateral Spectral-Aware Feature Pyramid Network (BS-FPN) is designed, which fuses the spectral context with object features at different scales by bi-directional sampling and enhances the features of small objects by Channel-wise Attention Modules (CAMs) and skipping connections.

3) Extensive experiments demonstrate the proposed approach achieves state-of-the-art results on the DIOR dataset and DOTA dataset, which prove the effectiveness of our proposed approach.

## 2. Methodology

There is a consensus that using context-assisted small object detection is effective in improving detection accuracy. It has been mainstreamed to use spatial operations to extract context, such as cross-scale fusion [7] and attention mechanism [6]. Recent studies on image classification [12] and domain generalization [16] have shown that simple operations on the frequency domain allow more effective extraction of long-term spatial dependencies, this is key to small object detection. The basic idea is that updating a value in the spectral domain affects all the original data comprehensively. Therefore, in this paper, we explore the extraction of frequency domain context to assist small object detection.

### 2.1 Overview of the proposed approach

Fig. 2 shows an overview of the proposed FFPF, which is based on the Faster R-CNN with FPN [7]. Compared with Faster R-CNN, there are mainly two novel components: Frequency-aware ResNet (F-ResNet) and Bilateral SpectralAware Feature Pyramid Network (BS-FPN). Specifically, compared to the original ResNet network, F-ResNet introduces a Fourier Unit (FU) based on spectral convolution as residuals after each block, which adds spectral global context through convolutional operations in the frequency domain. BS-FPN uses a bidirectional structure [17] to fuse the spectral global context with the small objects features and uses Channelwise Attention Modules (CAM) and skipping connections to enhance the small object features. In addition, the carafe operator [18] is employed to avoid the loss of information during the upsampling process.

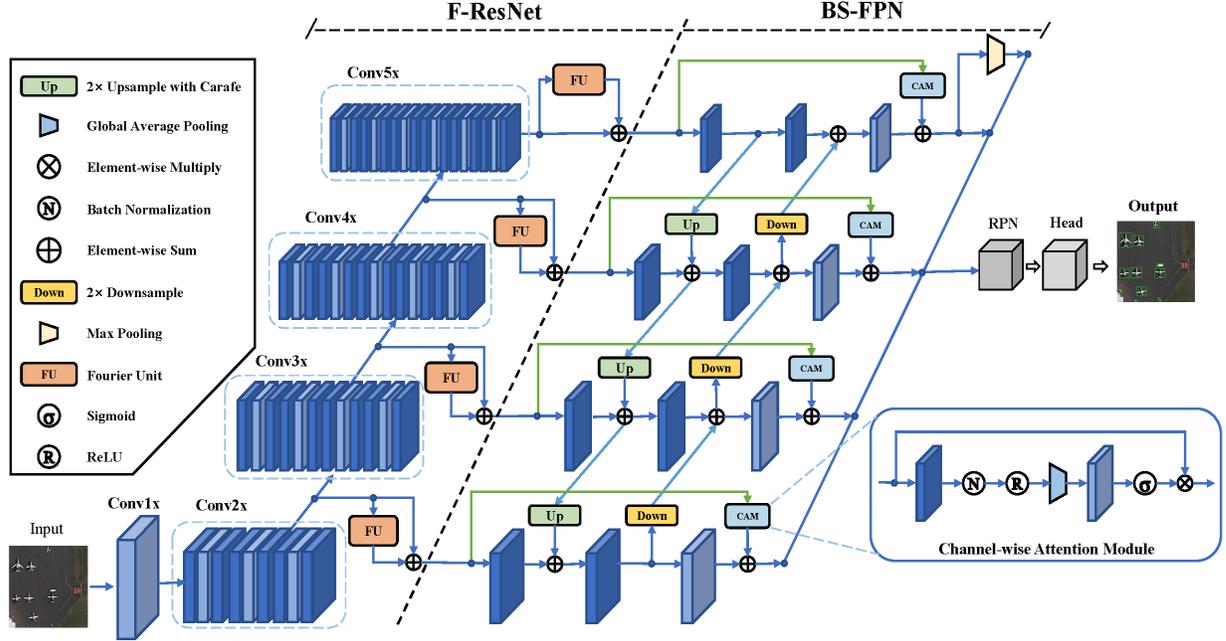

Fig. 2. Overview of the proposed approach.

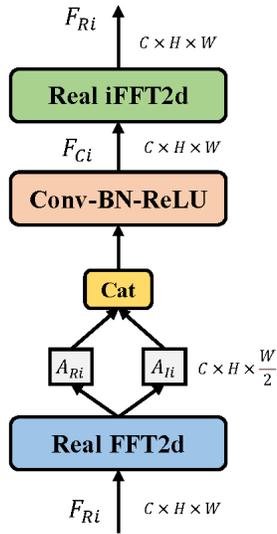

Fig. 3. Details of the Fourier Unit.

*2.2 Frequency-aware ResNet*

Frequency-aware ResNet (F-ResNet) consists of two parts: the vanilla ResNet network and the Fourier Unit (FU) residual layer. The basic idea of FU is to learn the interactions between spatial locations in the frequency domain, the details of which are shown in Fig. 3. According to the spectral convolution theorem in Fourier theory [14], updating a value in the spectral domain affects all the original data comprehensively [11].

Inspired by this, a Fourier Unit based on Fast Fourier Convolution is proposed, which uses frequency domain convolution to extract global context as supplementary information of small objects, thus assisting in the detection of small objects.

The details of F-ResNet are shown in Fig. 2. First, the feature maps extracted from the ResNet blocks are transformed from the original spatial domain to the spectral domain by the Fast Fourier Transform (FFT):

$$A_{Ri}, A_{Ii} = FFT(F_{Ri}) \quad (1)$$

where the real tensor $F_{Ri} \in \mathbb{R}^{C \times H \times W}$ denotes the output feature on $i$-th stage of ResNet, $A_{Ri}, A_{Ii} \in \mathbb{R}^{C \times H \times \frac{W}{2}}$ refer to the real and imaginary parts of the $i$-th feature map on the frequency domain, respectively.

Then the real and imaginary parts of the spectral feature map are concatenated, followed by the convolution operation in the frequency domain, resulting in a global update of the spectral data:

$$A_{Ci} = \text{Re}LU\left(BN\left(Conv\left(Cat\left(A'_{Ri}, A'_{Ii}\right)\right)\right)\right) \quad (2)$$

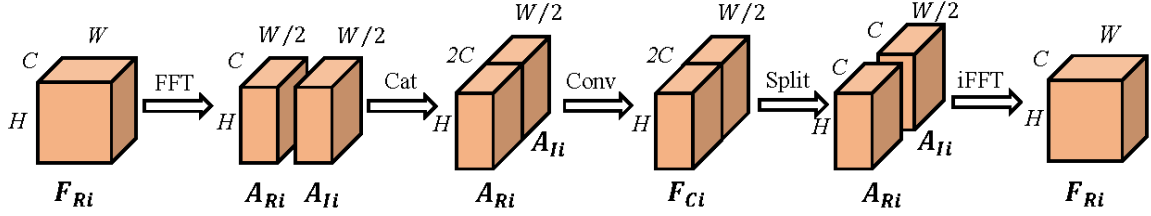

Fig. 4. Description of the flow of the feature map in the Fourier module. In which, "Cat" denotes the join operation along the channel dimension, "Conv" denotes the 1 × 1 convolution-batch normalization-ReLU operation, and "Split" denotes the segmentation operation along the channel dimension.

where $A_{Ci} \in \mathbb{C}^{2C \times H \times \frac{W}{2}}$ refers to the feature map of $i$-th layer, Conv denotes the $1 \times 1$ convolutional operation, and *Cat* refers to concatenation operations in the channel dimension. Then the spectral feature map is split into real and imaginary parts again, then transformed back into the spatial domain by inverse Fast Fourier Transform (iFFT).

$$A_{Ri} = iFFT(split(A_{Ci})) \quad (3)$$

where $A_{Ri} \in \mathbb{R}^{C \times H \times W}$ denotes the spatial feature map, iFFT denotes the inverse Fast Fourier Transform operation, and *split* denotes the splitting operation on the channel dimension. The above is the operation of FU where the spectral global context is obtained. The computational pipeline of FU is shown in Fig. 4. Finally, we fuse original features and the spectral global context by element-wise sum operation:

$$G_i = A_{Ri} + F_{Ri} \quad (4)$$

where $G_i$ is the final output on $i$-th global branch.

*2.3 Bilateral Spectral-Aware Feature Pyramid Network*

The shallow features in F-ResNet contain richer texture information, while the deep features contain more semantic information. In this paper, a Bilateral Spectral-aware FPN is proposed to obtain rich spectral-aware features by fully fusing the deep and shallow spectral context-aware features from FResNet. After obtaining the spectral context-aware features, BS-FPN first enriches the semantic features in the shallow layer by combining the semantic features extracted in the deep layer with the texture features extracted in the shallow layer through a top-down process. Then, through a bottomup shortcut, the small object features in the shallow layer are introduced into the deep features to avoid the small object features in the shallow layer lost in the long path. Finally, through CAM and skipping connections, the global spectralaware features from F-ResNet are enhanced and combined with the fused features to avoid the introduction of noise when fused with features in different layers.

The details of the BS-FPN are shown to the right of the black dashed line in Fig. 2. First, the deep feature map is upsampled and then combined with the shallow feature map to construct the top-to-bottom path:

$$U_i = Conv(G_i + Upsample(U_{i+1})) \quad (5)$$

where $U_i \in \mathbb{R}^{C \times H \times W}$ denotes the $i$-th feature, *Conv* denotes the $1 \times 1$ convolution operation, and *Upsample* is the carafe [18] upsampling operation.

Then, the merged shallow feature map is downsampled and then combined with the deep feature map:

$$B_{i+1} = Conv(U_{i+1} + Downsample(B_i)) \quad (6)$$

where $B_i \in \mathbb{R}^{C \times H \times W}$ denotes the $i$-th feature map, *Conv* is a 1 × 1 convolutional operation, and *Downsample* denotes a 3 × 3 convolutional operation with stride size 2.

In addition, CAMs are used to enhance the frequency domain context-aware features of F-ResNet. First, global information about the features in the current layer is obtained through a Global Averaging Pooling (GAP) operation:

$$Z_i = \frac{1}{W \times H} \sum_{j=1}^{W} \sum_{k=1}^{H} (T_1 G_i) \quad (7)$$

where $Z_i \in \mathbb{R}^{C \times 1 \times 1}$ denotes the output of the Global Average Pooling layer. $T_1$ denotes the convolution-batch normalization-ReLU layer, where the

Table 1. Comparison with state-of-the-art methods for the DIOR dataset. We mark the results of the best results by BOLD text.

| Model | YOLOv3[19] | FRCNN[3] | PANet[17] | FPN[7] | MRCNN[22] | RetinaNet[23] | FPN[7] | Carafe[18] | CSFF[24] | GCF[25] | Ours |
|---|---|---|---|---|---|---|---|---|---|---|---|
| Backbone | Darknet53 | VGG16[20] | ResNet50[21] | ResNet50 | ResNet50 | ResNet101[21] | ResNet101 | ResNet101 | ResNet101 | ResNet101 | ResNet50 |
| C1 | **72.2** | 53.6 | 61.9 | 64.4 | 53.8 | 53.3 | 60.2 | 58.9 | 57.2 | 62.8 | 65.5 |
| C2 | 29.2 | 49.3 | 70.4 | 81.1 | 72.3 | 77.0 | 83.4 | 83.7 | 79.6 | 86.5 | **86.7** |
| C3 | 74.0 | 78.8 | 71.0 | 72.7 | 63.2 | 69.3 | 73.8 | 77.8 | 70.1 | 74.8 | **79.4** |
| C4 | 78.6 | 66.2 | 80.4 | 89.1 | 81.0 | 85.0 | 88.7 | 88.9 | 87.4 | **89.2** | 89.0 |
| C5 | 31.2 | 28.0 | 38.9 | 47.2 | 38.7 | 44.1 | 49.0 | **50.6** | 46.1 | 49.2 | 50.3 |
| C6 | 69.7 | 70.9 | 72.5 | **80.1** | 72.6 | 73.2 | 78.9 | 79.1 | 76.6 | 76.6 | 79.2 |
| C7 | 26.9 | 62.3 | 56.6 | 62.3 | 55.9 | 62.4 | 66.7 | 72.7 | 62.7 | 72.5 | **73.3** |
| C8 | 48.6 | 69.0 | 68.4 | 81.5 | 71.6 | 78.6 | 85.4 | 82.8 | 82.6 | 85.7 | **87.6** |
| C9 | 54.4 | 55.2 | 60.0 | 70.6 | 67.0 | 62.8 | 71.3 | 72.7 | 73.2 | **75.1** | 73.6 |
| C10 | 31.1 | 68.0 | 69.0 | 79.1 | 73.0 | 78.6 | 81.5 | 82.8 | 78.2 | 81.3 | **83.5** |
| C11 | 61.1 | 56.9 | 74.6 | 82.4 | 75.8 | 76.6 | 82.8 | 84.3 | 81.6 | 83.3 | **85.1** |
| C12 | 44.9 | 50.2 | 41.6 | 54.2 | 44.2 | 49.9 | 54.7 | 55.8 | 50.7 | **60.2** | 57.3 |
| C13 | 49.7 | 50.1 | 55.8 | 61.7 | 56.5 | 59.6 | 62.4 | 62.4 | 59.5 | 62.7 | **63.5** |
| C14 | **87.4** | 27.7 | 71.7 | 74.3 | 71.9 | 71.1 | 73.3 | 74.3 | 73.3 | 72.7 | 74.1 |
| C15 | 70.6 | 73.0 | 72.9 | 74.5 | 58.6 | 68.4 | 77.3 | 75.2 | 63.4 | 77.3 | **78.4** |
| C16 | **68.7** | 39.8 | 62.3 | 58.8 | 53.6 | 45.8 | 59.4 | 59.0 | 58.5 | 61.9 | 59.3 |
| C17 | 87.3 | 75.2 | 81.2 | 87.6 | 81.1 | 81.3 | 87.5 | 88.7 | 85.9 | 88.0 | **88.6** |
| C18 | 29.4 | 38.6 | 54.6 | 62.9 | 54.0 | 55.2 | 65.0 | 70.4 | 61.9 | 69.9 | **71.0** |
| C19 | **48.3** | 23.6 | 48.2 | 42.1 | 43.1 | 44.4 | 42.2 | 43.6 | 42.9 | 47.0 | 43.3 |
| C20 | 78.7 | 45.4 | 86.7 | 86.4 | 81.1 | 85.5 | 85.1 | 86.8 | 86.9 | **89.7** | 87.4 |
| mAP | 57.1 | 54.1 | 63.8 | 70.6 | 63.5 | 66.1 | 71.4 | 72.8 | 68.0 | 73.3 | **73.8** |

convolutional kernel is 3. Then the activation operation is performed on the attention vector:

$$S_i = \sigma(T_2 Z_i) \qquad (8)$$

in which $S_i \in \mathbb{R}^{C \times 1 \times 1}$ refers to the attention vector, $\sigma$ represents the sigmoid function, and T2 denotes the 1×1 convolution. Finally, the features from CAM are combined with features from bilateral sampling at the same layer using skip connections:

$$L_i = B_i + S_i \times G_i \qquad (8)$$

where $L_i$ is the $i$-th output of the BS-FPN, $S_i$ is the $i$-th CAM with the details shown in Fig. 2, and × is the element-wise multiply.

## 3. Experiments

*3.1 Dataset*

In this paper, all experiments are conducted on two largescale datasets named DIOR [15] and DOTA [26], which are public available large datasets proposed for remote sensing image object detection.

DIOR dataset contains 23,463 remote sensing images, 192,472 annotated instances, with 20 object categories: Airplane (C1), Airport (C2), Baseball field (C3), Basketball court (C4), Bridge (C5), Chimney (C6), Dam (C7), Expressway service area (C8), Expressway toll station (C9), Golf field (C10), Ground track field (C11), Harbor (C12), Overpass (C13), Ship (C14), Stadium (C15), Storage tank (C16), Tennis court (C17), Train station (C18), Vehicle (C19), and Windmill (C20). DIOR dataset is sourced from Google Earth with an image size of 800 × 800 pixels and the spatial resolutions ranging from 0.5 to 30 m, where each instance is labeled using a horizontal bounding box. In the experiments of this paper, we train on the Trainval subset of the DIOR dataset and test on the test subset.

DOTA is a multi-scale optical remote sensing dataset with images mainly from Google Earth, the JL-1 and GF-2 satellites of the China Resources Satellite Data and Applications Centre. DOTA dataset contains 2806 aerial images, each of which is from about 800×800 to 4000×4000 pixels in size. This paper uses the HBB annotation from the DOTA-v1.5 dataset, which contains 400,000 optically remote sensed objects of various scales, orientations and shapes in 16 categories (e.g. container crane (CC), baseball field (BD), basketball court (BC), bridge (B), surface runway (GTF), harbour (HB), helicopter (H), large Vehicle (LV), Aircraft (P), Roundabout (R), Ship (S), Small Vehicle (SV), Football Field (SBF), Storage Tank (ST), Swimming Pool (SP) and Tennis Court (TC)). The images will be cropped into small image patches with an overlap of 200 through a sliding window of 800 × 800 size..

*3.2 Implementation Details*

The experiments were implemented using Pytorch 1.7.0 and cuda11.0, based on MMDetection toolbox. The baseline in this paper is the Faster R-CNN with FPN, with two images computed on a single GPU. The pre-trained weights on ImageNet are used and all new layers are initialized using kaiming normal. The total epoch is 12 and the initial learning rate is 0.01, decreasing by 0.1 after the 8th and 11th epochs, respectively, as the original settings in the official mmdetection codebase during training. No data augmentation is used in this paper.

*3.3 Comparison With State-of-the-Art approaches*

We conduct experiments on the DIOR dataset and DOTA dataset to compare the performance of our approach with several state-of-the-art approaches. The results are shown in Tables 1-2.

Table 1 shows the detection accuracy of all approaches. Among them, the backbone of YOLOv3 [19] is Darknet53, the backbone of Faster R-CNN [3] is VGG16 [20], the backbone of PANet [17], Mask R-CNN [22] and FPN [7] is ResNet50 [21], and all other approaches are implemented by using ResNet101 [21] as the backbone.

Table 2 Comparison with classical methods for the DOTA dataset.

| model | mAP |
|---|---|
| Faster-RCNN | 56.6 |
| YOLOv3 | 55.1 |
| RetinaNet | 42.4 |
| Ours | **57.6** |

It can be seen that our proposed approach achieve the highest mAP and outperformed all comparison approaches in half of the categories, which verifies the effectiveness of our approach.

Since the splitting methods in the DOTA dataset are not uniform, we only compared the classical open-source object detection approaches on the DOTA dataset, as showed in Table 2. The DOTA dataset contains a large number of small objects, so it can represent scenes with many small objects in remote sensing. As can be seen, the detection performance of this paper's FFPF is always the best compared to other peer approaches. The results show that the FFPF of this paper can effectively handle scenes with many small objects.

*3.4 Ablation Studies*

To further demonstrate the effectiveness of the proposed F-ResNet and BS-FPN, four sets of ablation studies are designed on the DIOR dataset. The results are shown in Table 3.

1) F-ResNet: When the Fourier Unit is added to the original backbone, mAP increases from 71.4% to 73.4%. Moreover, the Fourier Unit is an inexpensive module, which introduces only a few additional parameters.

2) BS-FPN: Compared with the baseline network (i.e., FPN), the performance of BS-FPN in terms of mAP is 1.7% higher than that of baseline. It demonstrates that the BS-FPN can obtain more context than the vanilla FPN, resulting in better performance on final detection.

Table 3 Ablation studies on the DIOR dataset.

| baseline | F-ResNet | BS-FPN | mAP |
|---|---|---|---|
| √ | | | 71.4 |
| √ | √ | | 73.4 |
| √ | | √ | 73.1 |
| √ | √ | √ | **73.8** |

Table II shows that each component (i.e., F-ResNet and BS-FPN) brings additional significant gain. The FFPF that jointly considers the F-ResNet and BS-FPN provides the best performance (73.8%), which shows that the proposed FFPF is indeed effective for remote sensing object detection task.

*3.5 Qualitative Analysis*

For a more visual qualitative analysis of our approach, the visualisation results are compared on the DIOR dataset and the DOTA dataset.

Fig. 5 shows some examples of FFPF detection results on the DIOR dataset, including the detection of objects of different sizes and scales such as aircraft, windmills, oil tanks, ground athletic fields and vehicles in a variety of scenarios. It can be seen that the FFPF in this paper has a good detection result for different objects in various scenarios, which demonstrates the robustness of the FFPF.

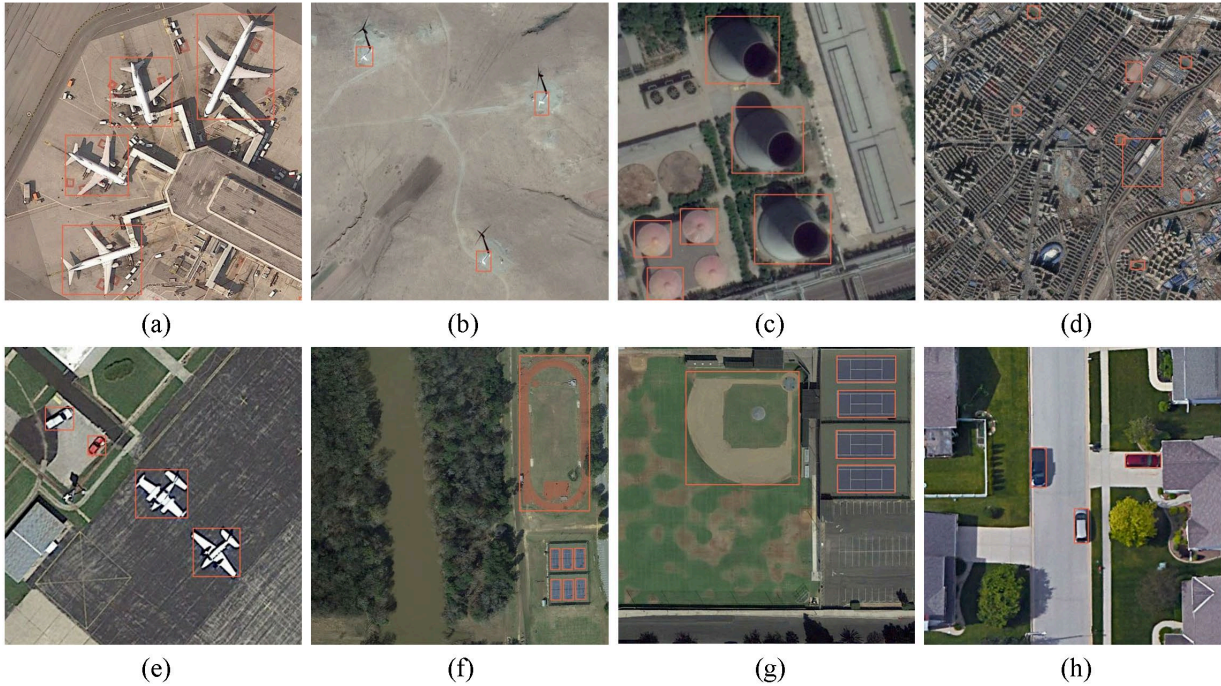

Fig. 5. Examples of detection results for FFPF on the DIOR dataset. (a) Aircraft. (b) windmills. (c) storage tanks. (d) Ground athletic field. (e) Aircraft and vehicles. (f) Ground athletic field and baseball field. (g) Stadiums and baseball fields. (h) Vehicles.

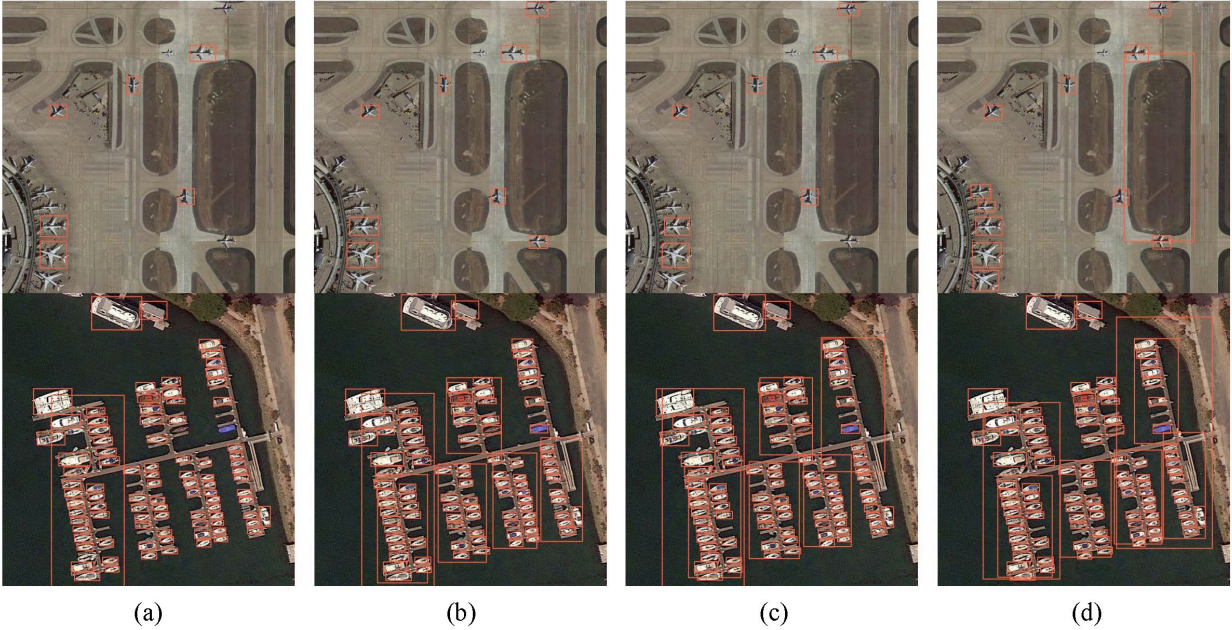

Fig. 6. Examples of detection results on the DIOR dataset. (a) Original Faster-RCNN. (b) Faster-RCNN + F-ResNet. (c) Faster-RCNN + BS-FPN. (d) FFPF.

The partial visualisation of the ablation experiment on the DIOR dataset is shown in Fig. 6. It can be seen that the introduction of the F-ResNet and the BS-FPN modules result in a lower miss detection rate compared to the original Faster R-CNN. When both modules are used simultaneously, our FFPF achieves an optimal detection result. This again demonstrates the effectiveness of our method in detecting remote sensing objects.

Fig. 7 shows the results of comparing our FFPF with the three classical approaches of Original Faster-RCNN, YOLOv3 and RetinaNet. Fig. 7 presents a scene with a ship moored in a harbour. The relative size of the ship is small and contains too few features, resulting in inaccurate detection. It can be seen that the YOLOv3 and RetinaNet networks have a higher miss detection rate and are less competent in detecting smaller objects, while Faster-RCNN has a better detection capability, but there are still some undetected objects. And the FFPF in this paper achieves the best detection results. This demonstrates our approaches' effectiveness in detecting remote sensing objects.

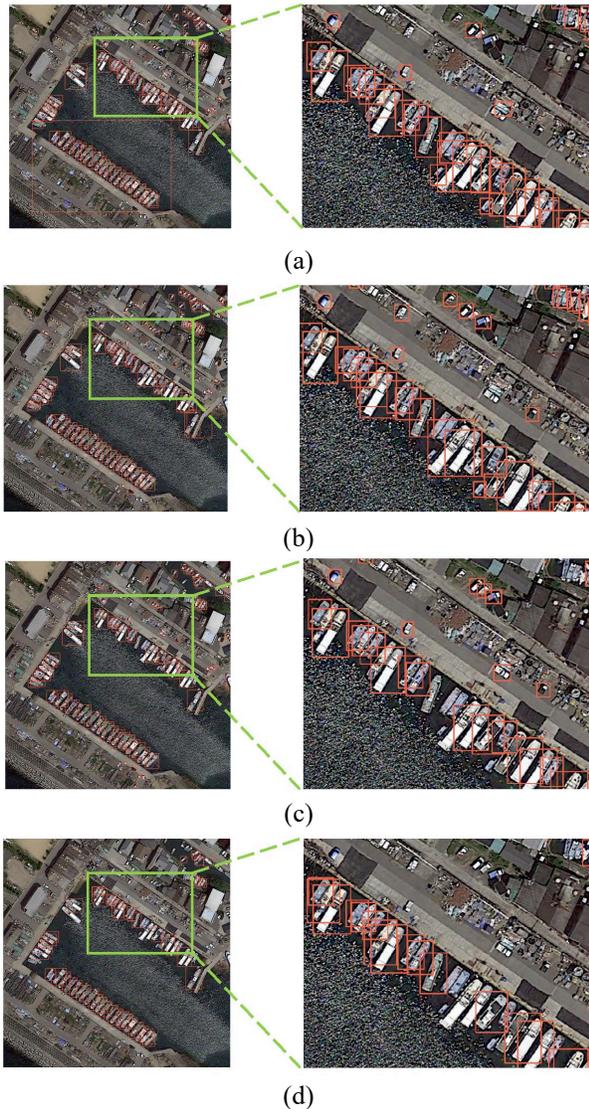

Fig. 7. Example of detection results on the DOTA dataset compared with the classical detection approaches. (a) FFPF. (b) Original Faster-RCNN. (c) YOLOv3. (d) RetinaNet.

In conclusion, from the qualitative analysis, it can be further demonstrated that our FFPF outperforms other approaches in terms of detection accuracy.

## 4. Conclusions

In this paper, we explore how to obtain the context simply and efficiently through frequency domain operations to improve the accuracy of remote sensing object detection. Specifically, we propose a Frequency-aware Feature Pyramid Framework, which consists of two main components: F-ResNet and BS-FPN. F-ResNet perceives the spectral context information to extract richer features of small objects. While BS-FPN model the association of object features at different scales. Extensive experiments demonstrate the effectiveness of FFPF. To the best of our knowledge, this is the first work to employ frequency domain convolution in remote sensing object detection. Nevertheless, there are some shortcomings in our approach. Specifically, the frequency-domain convolution module designed in this paper is relatively simple, and the investigation of the frequency-domain convolution operation is not yet deep enough. In the future, we will further investigate how to better apply operations in the frequency domain on remote sensing object detection.

**Acknowledgements**

The study was realized with the support of the China Scholarship Council (Grant No. 202106210056) and was supported in part by the program «Best International Grant for PhD» of Peter the Great St.Petersburg Polytechnic University".